\documentclass[letterpaper, 10 pt, conference]{ieeeconf}  %

\IEEEoverridecommandlockouts                              %

\usepackage{graphics} %
\usepackage{epsfig} %
\usepackage{mathptmx} %
\usepackage{times} %
\usepackage{amsmath} %
\usepackage{amssymb}  %
\usepackage[T1]{fontenc}
\usepackage{tabu}
\usepackage{pifont}%
\usepackage{wrapfig}

\usepackage[font=scriptsize]{caption}
\newcommand{\cmark}{\ding{51}}%
\newcommand{\imark}{ \ding{55}}

\usepackage{array}
\usepackage{makecell}

\usepackage{tabu}
\usepackage{xcolor}
\newcommand{\specialcell}[2][c]{%
\begin{tabular}[#1]{@{}c@{}}#2\end{tabular}}

\usepackage[colorlinks = true,
            linkcolor = blue,
            urlcolor  = blue,
            citecolor = blue,
            anchorcolor = blue]{hyperref}
\newcommand{\MYhref}[3][blue]{\href{#2}{\color{#1}{#3}}}%

\title{\LARGE \bf
Bridge Data: Boosting Generalization of Robotic Skills with Cross-Domain Datasets
}

\author{Frederik Ebert*$^{1}$, Yanlai Yang*$^{1}$, Karl Schmeckpeper$^{3}$, Bernadette Bucher$^{3}$, Georgios Georgakis$^{3}$, \\  Kostas Daniilidis$^{3}$, Chelsea Finn$^{2}$, Sergey Levine$^{1}$
\thanks{* denotes equal contribution}
\thanks{$^{1}$ University of California Berkeley, $^{2}$ Stanford University, $^{3}$ University of Pennsylvania}
}

\begin{document}

\maketitle
\thispagestyle{empty}
\pagestyle{empty}

\begin{abstract}

Robot learning holds the promise of learning policies that generalize broadly. However, such generalization requires sufficiently diverse datasets of the task of interest, which can be prohibitively expensive to collect.
In other fields, such as computer vision, it is common to utilize shared, reusable datasets, such as ImageNet, to overcome this challenge, but this has proven difficult in robotics. In this paper, we ask: what would it take to enable practical data reuse in robotics for end-to-end skill learning?
We hypothesize that the key is to use datasets with multiple tasks and multiple domains, such that a new user that wants to train their robot to perform a new task in a new domain can include this dataset in their training process and benefit from cross-task and cross-domain generalization. To evaluate this hypothesis, we collect a large multi-domain and multi-task dataset, with 7,200 demonstrations constituting 71 tasks across 10 environments, and empirically study how this data can improve the learning of new tasks in new environments.
We find that jointly training with the proposed dataset and 50 demonstrations of a never-before-seen task in a new domain on average leads to a 2x improvement in success rate compared to using target domain data alone. We also find that data for only a few tasks in a new domain can bridge the domain gap and make it possible for a robot to perform a variety of prior tasks that were only seen in other domains. These results suggest that reusing diverse multi-task and multi-domain datasets, including our open-source dataset, may pave the way for broader robot generalization, eliminating the need to re-collect data for each new robot learning project.

\end{abstract}

\section{Introduction}
\label{sec:introduction}

Humans and animals can generalize a learned skill to a wide variety of contexts without needing to relearn the skill every time.  Endowing robots with the same capability would be a significant advance toward making robots more applicable to a range of real-world settings.
However, the prevailing paradigm of robot learning is to repeat data collection and policy training from scratch for every new task and environment. Learning policies in isolation not only increases the costs of data collection, but also limits the policy's scope of generalization.

In other fields, such as computer vision \cite{krizhevsky2012imagenet} and natural language processing (NLP) \cite{devlin2018bert}, utilizing large, diverse datasets has shown considerable success in enabling generalization to new problems or domains with a small amount of data (e.g., via pretraining and finetuning). 
However, in robotics, datasets are usually collected with a specific robotic platform and domain in mind, typically by the same researcher who intends to use that dataset. What would it take to make datasets reusable in robotics in the same way as large supervised datasets are reused (e.g., ImageNet~\cite{deng2009imagenet})? Each end-user of such a dataset might want their robot to learn a different task, which would be situated in a different domain (e.g., a different laboratory, home, etc.). It is currently an open question whether such reuse is feasible in robotics, and we posit that any such dataset would need to cover both multiple different tasks and multiple different domains. To this end, the aim of our paper is to investigate the degree to which such a multi-task and multi-domain dataset, which we refer to as a \emph{bridge} dataset, can enable a new robot in a new domain (which was not seen in the bridge data) to more effectively generalize when learning a new task (which was also not seen in the bridge data), as well as to transfer tasks from the bridge data to the target domain. We also propose a new dataset that enables this goal in the context of kitchen-themed tasks with a low-cost robotic arm and is intended to be reused by other researchers.

\begin{figure}
  \includegraphics[width=1\linewidth, bb=0 0 580 290]{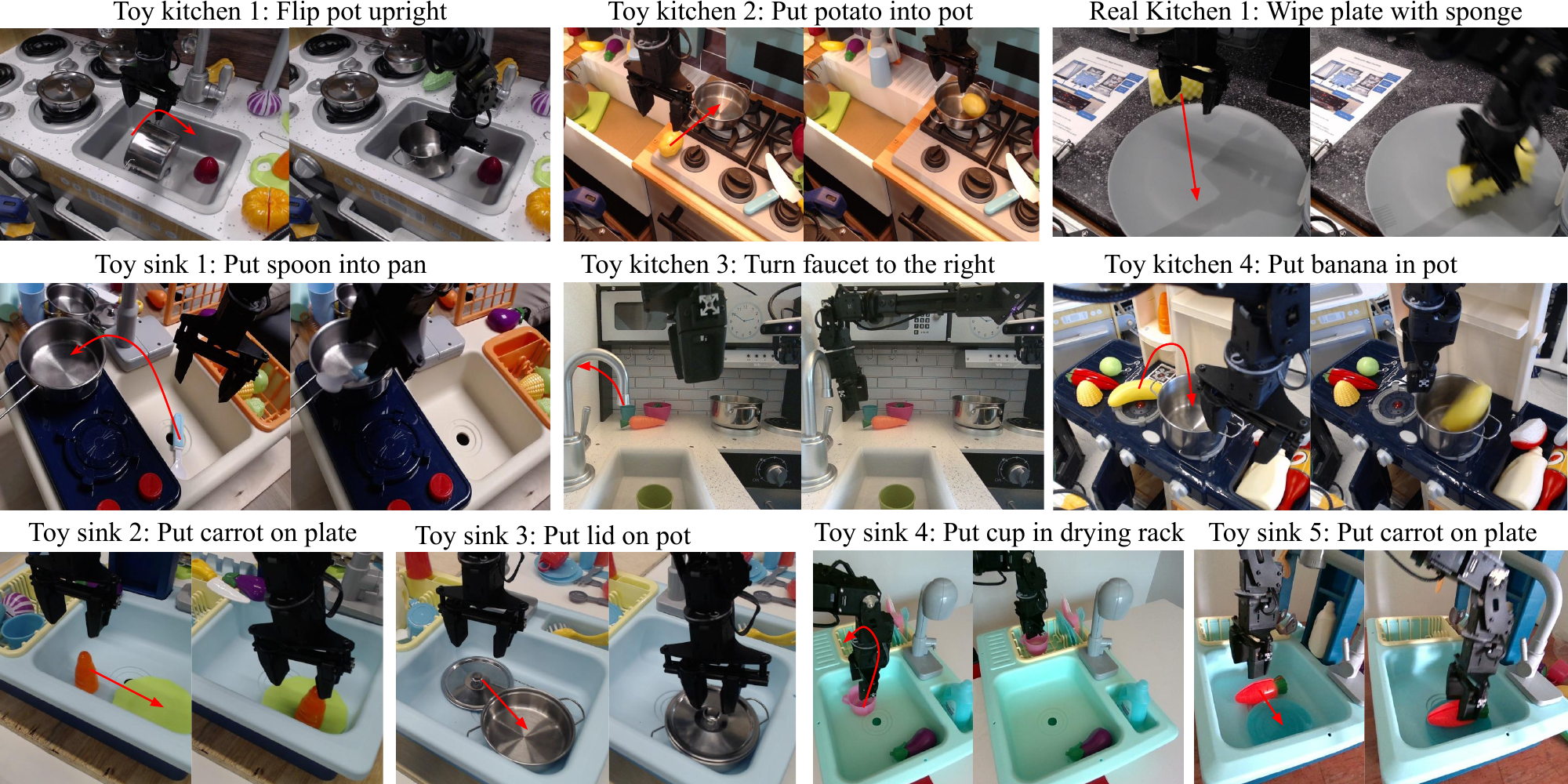}
  \caption{Illustration of our bridge dataset. The dataset includes demonstrations in 10 environments (4 toy kitchens and 5 toy sinks and 1 real kitchen), collected using a WidowX250 robot controlled via an Oculus Quest2 VR device, and consists of 7200 demonstrations. The red arrows indicate the desired movement of the target object.}
  \label{fig:dataset}
\end{figure}

The notion that multi-task data can speed up learning or improve generalization has been studied in many prior works~\cite{yu2020gradient, kalashnikov2021mt}. However, unlike this paper, the focus in these prior works, as we discuss in Section~\ref{sec:related_work}, is not on enabling new users to quickly train generalizable skills in a new setting or domain, but rather to utilize multi-task learning to lower the data requirements of acquiring a pre-defined set of tasks. More closely related to our work, RoboNet~\cite{dasari2019robonet} contains data from multiple robots and domains, but this data is collected using random motions, and does not provide examples of multiple different tasks that can be used for more complex task-directed manipulation. We discuss other datasets in Section~\ref{sec:related_work}; but in summary, no existing dataset covers \emph{both} multiple tasks \emph{and} multiple domains in a way that is suitable to study our central hypothesis: can prior data be used to improve the generalization of \emph{new} tasks in \emph{new} domains? We will call this the \emph{bridge data hypothesis}. We believe this is a critical requirement for effective data reuse in robotics, where different labs and researchers can all bootstrap from the same shared datasets. To study this, 
we collected a new multi-domain manipulation dataset with 7,200 demonstrations of 71 distinct and semantically meaningful tasks, themed around household tasks in kitchen environments. The data was collected across 10 distinct ``toy'' kitchens, as shown in Figure~\ref{fig:dataset}.
This data is suitable for imitation learning, which is the focus of our work, though it could also be repurposed for offline RL and other algorithms in the future.
We present our new dataset, and then use it to evaluate the bridge data hypothesis that is stated above, using three types of transfer scenarios: \textbf{(1)} When the user needs to train an existing task in a new domain, does the inclusion of bridge data boost performance? This roughly corresponds to a standard domain adaptation setting. \textbf{(2)} After the user has collected some data for a few tasks in a new domain, can their robot then perform other tasks that were \emph{not} seen in the new domain, but are only present in the bridge data (i.e., can it ``import'' tasks from the bridge data)? \textbf{(3)} When the user collects some data in a new domain for a task that was \emph{not} seen in the bridge data, can the performance and generalization of this task be boosted by including the bridge data in training? Scenario \textbf{(3)} directly evaluates our central hypothesis, while the other scenarios illustrate other potential uses for bridge data.

The main contributions of our work consist of an empirical evaluation of the bridge data hypothesis and a practical example of a bridge dataset with 7,200 demonstrations for 71 tasks in 10 environments, which we have released publicly on the project website.\footnote{\url{https://sites.google.com/view/bridgedata}}
To the best of our knowledge, our work is also the first to demonstrate transfer scenarios \textbf{(2)} and \textbf{(3)} above. This is significant, because \textbf{(2)} provides users with a low-cost way to ``import'' all of the skills in the bridge dataset into their own domain with just a small number of demonstrations in their domain, while \textbf{(3)} provides for a way to boost the performance of an entirely new skill with previously collected reusable bridge data. Our results suggest that accumulating and reusing diverse multi-task and multi-domain datasets, at least when all data is collected with the same type of robot, may make it possible for researchers to endow robots with generalizable skills using only a modest amount of in-domain data for their desired task.
\section{Related Work}

\label{sec:related_work}

While most prior work on deep visuomotor learning trains a single task in a single domain \cite{finn2017one, duan2017one, ho2016generative, yu2018one, liu2018imitation, sermanet2018time, ghadirzadeh2020human, tian2020model,zhang2018deep}, our goal is not to develop better learning methods, but rather to illustrate how generic multi-domain, multi-task datasets can be used with existing algorithms to boost the generalization of \emph{new} tasks in \emph{new} domains.
Prior work on multi-task reinforcement learning \cite{kalashnikov2021mt} has shown that data from other tasks can boost generalization of new tasks, however this study is carried out in a \emph{single} domain.

Existing robot learning datasets do not exhibit the right properties for boosting the generalization of new tasks in \emph{new} domains or zero-shot transferring skills from the prior dataset to a target domain. We provide an overview of the most related datasets in \autoref{tab:dataset_comparison}.
Most existing robot datasets, such as MIME~\cite{sharma2018multiple}, DAML \cite{yu2018one}, RoboTurk~\cite{mandlekar2018roboturk,mandlekar2019scaling}, and many others \cite{pinto2016supersizing,finn2017deep,levine2018learning,kalashnikov2018scalable,ebert2018visual, zeng2020tossingbot, kalashnikov2021mt} only feature a single domain, making them difficult to use for boosting the generalization in \emph{other} domains. Merging multiple existing datasets into one multi-domain dataset is difficult due to inconsistencies in data collection protocols, time discretization, robot morphologies, and sensors. Learning from multiple robots has been studied with RoboNet~\cite{dasari2019robonet}, which provides a dataset with 7 different robots in different domains. Here the data is generated with random motions which do not produce semantically meaningful tasks. This limits task complexity to pushing and basic grasping, and makes the data poorly suited for imitation learning.

\begin{figure}
    \scriptsize
    \centering
    \begin{tabular}{|l|l|l|l|l|}
    \hline
    Dataset &  \# Tasks  & \# Trajec.  & \# Domains  & \thead{suitable\\for BC/IL}\\ \hline
    DAML \cite{yu2018one} & 3 & 2.9k & 1 & \cmark \\
    MIME \cite{sharma2018multiple} & 22 & 8.2k & 1 & \cmark \\ 
    RoboNet \cite{dasari2019robonet} & N/A & 162k & 7 & \imark  \\
    RoboTurk \cite{mandlekar2018roboturk, mandlekar2019scaling} & 3 & 2.1k & 1 & \cmark \\ 
    Vis. Imit. Made \cite{young2020visual}  & 2 & 2k & 50 & \cmark \\
    \hline
    \textbf{Ours} & \textbf{71} & 7.2k & 10 & \cmark \\
    \hline
    
    \end{tabular}
    \caption{Comparison of our dataset and prior works. Our dataset has by far the most tasks, and is the only dataset with more than 2 tasks that has many domains. This is critical for evaluating the bridge data hypothesis.}
    \label{tab:dataset_comparison}
\end{figure}

Some prior works have also used datasets collected by humans without a robot, across multiple domains. For example, Young et al.~\cite{young2020visual} presents results on data across many more domains than our bridge data, collected via a hand-held gripper, but only presents two grasping tasks.

%
\section{Bridge Datasets}
\label{sec:method}

In this section, we describe the basic principles behind bridge datasets and how they can be used to boost generalization. Then, we present a description of the specific bridge dataset that we collected using teleoperation of a low-cost robotic arm for a range of kitchen-themed manipulation tasks. We use the term \emph{bridge dataset} to refer to a large and diverse dataset of robotic behaviors collected in a range of settings (e.g., different viewpoints, lighting conditions, objects, and scenes), for a range of \emph{different} tasks, so as to make it possible to ``bridge'' gaps in the generalization that arise when the user provides a small to medium amount of data in their specific target domain. We define the term ``target domain'' to refer to the environment where the robot must perform the desired task. This target domain is distinct from \emph{any} of the settings seen in the bridge dataset: the intent is for the same large bridge dataset to be used by all users for whichever target domain they require.

\begin{figure}
\centering
 \includegraphics[width=0.7\linewidth]{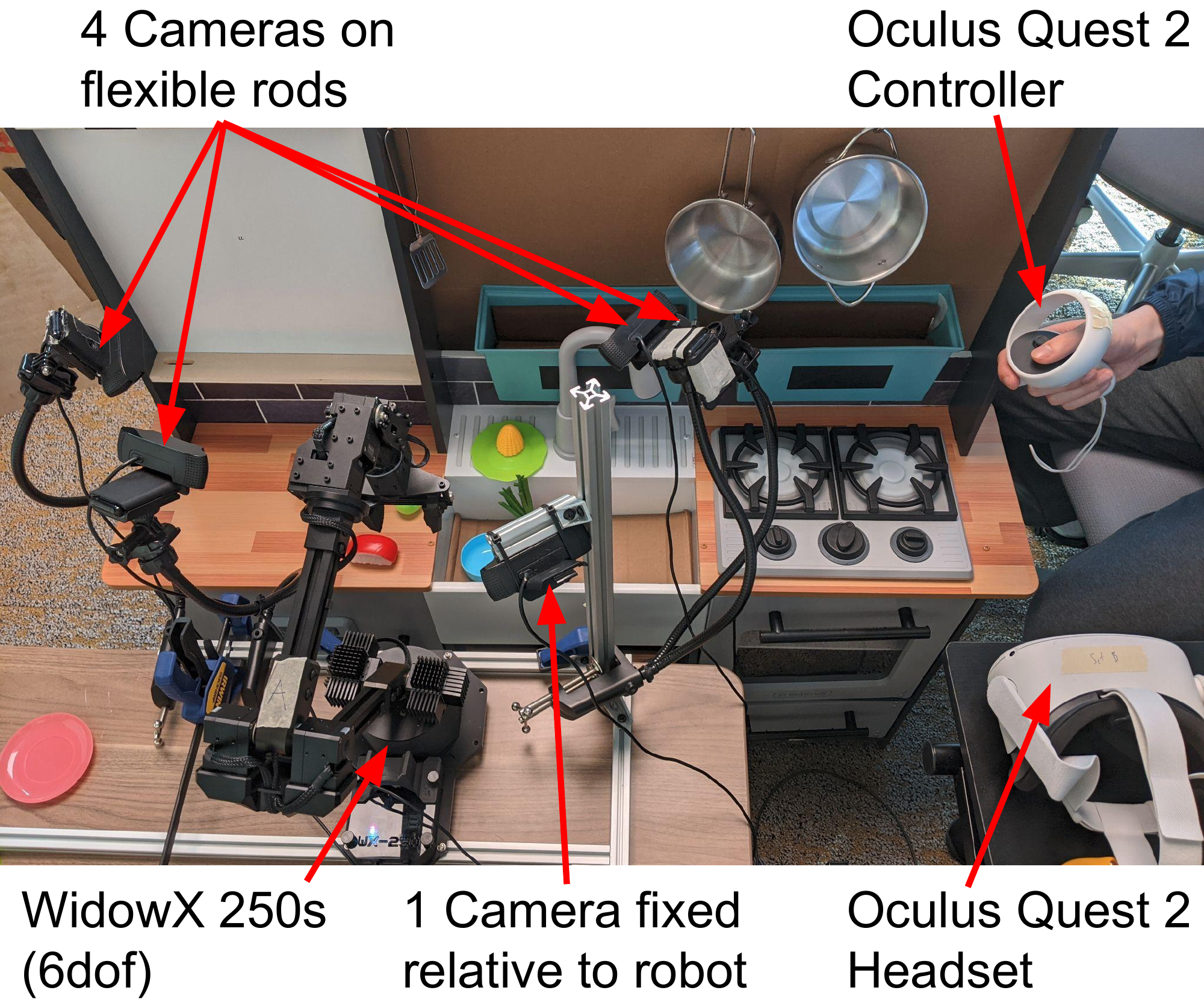}
  \caption{Demonstration data collection setup using VR Headset. The scene is captured by 5 cameras simultaneously. While one of the cameras is fixed, the others are mounted on flexible rods.}
  \label{fig:system_overview}
\end{figure}

\begin{figure}
\centering
  \includegraphics[width=1.0\linewidth]{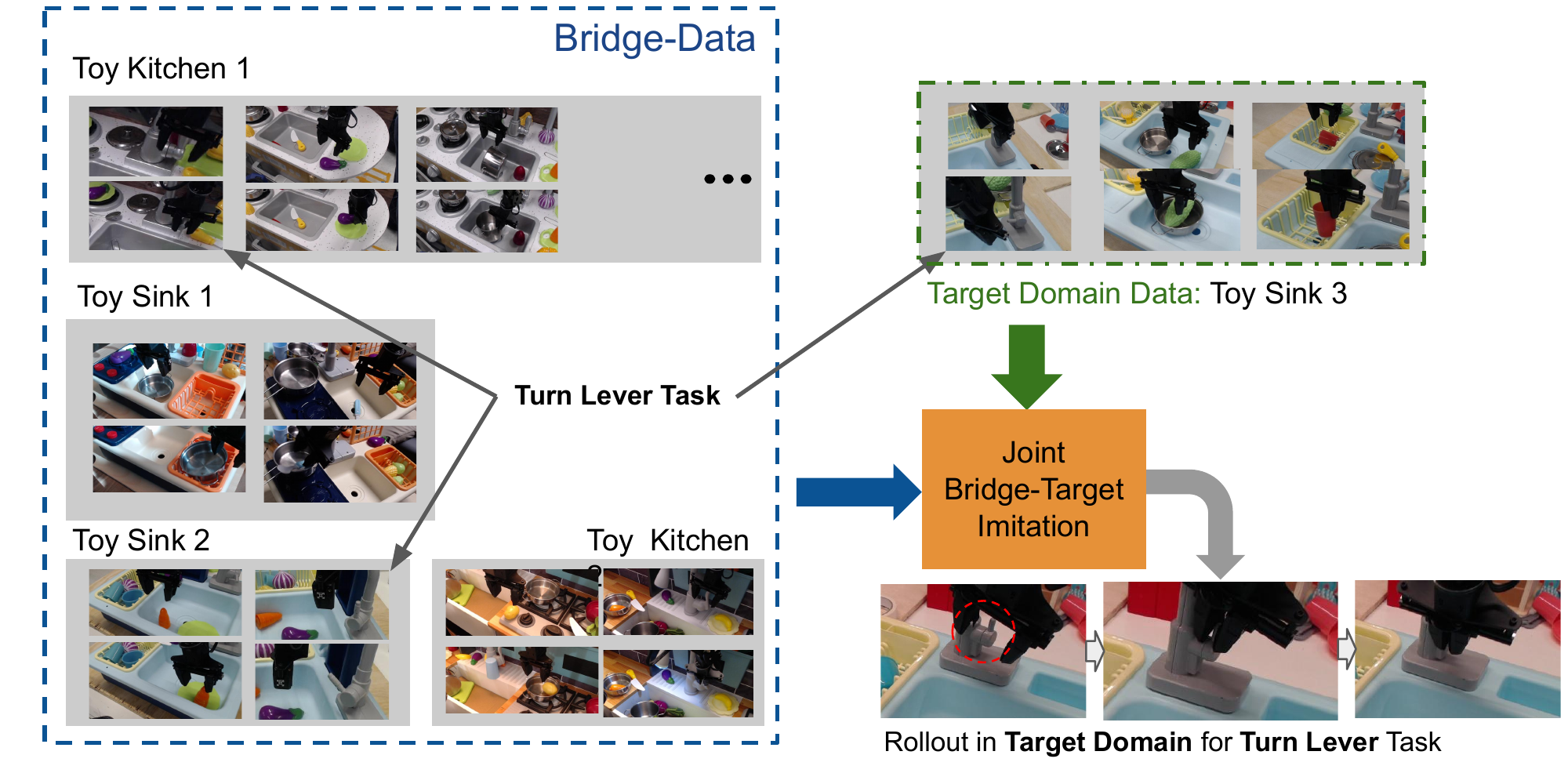}
  \caption{\textbf{Scenario (1): transfer with matching behaviors}. In this setting, bridge data is used to improve the performance and generalization of tasks in the target domain for which the user has collected some amount of data. These tasks must also be present in the bridge data. In this example, the user demonstrates the ``turn lever,'' ``squash into pot,'' and ``flip cup'' tasks in the target domain, and these tasks are also present in several domains in the bridge data. After including the bridge data in training, the performance and generalization of these tasks in the target is significantly higher.}
  \label{fig:matching_behaviors}
\end{figure}
\subsection{Boosting Generalization via Bridge Datasets}
\label{sec:boosting}
We consider three types of generalization in our experiments, though other modes may also be feasible:

\noindent \textbf{(1) Transfer with matching behaviors}, where the user collects some small amount of data in their target domain for tasks that are also present in the bridge data (e.g., around 50 demos per task), and uses the bridge data to boost the performance and generalization of these tasks. We illustrate this scenario in Figure~\ref{fig:matching_behaviors}.
This scenario is the most conventional, and resembles domain adaptation in computer vision, but it is also the most limiting, since it requires the user's desired tasks to be present in the bridge data.
However, as we will show, bridge data can enable very significant performance and generalization boosts in this setting.

\begin{figure}
\centering
  \includegraphics[width=\linewidth]{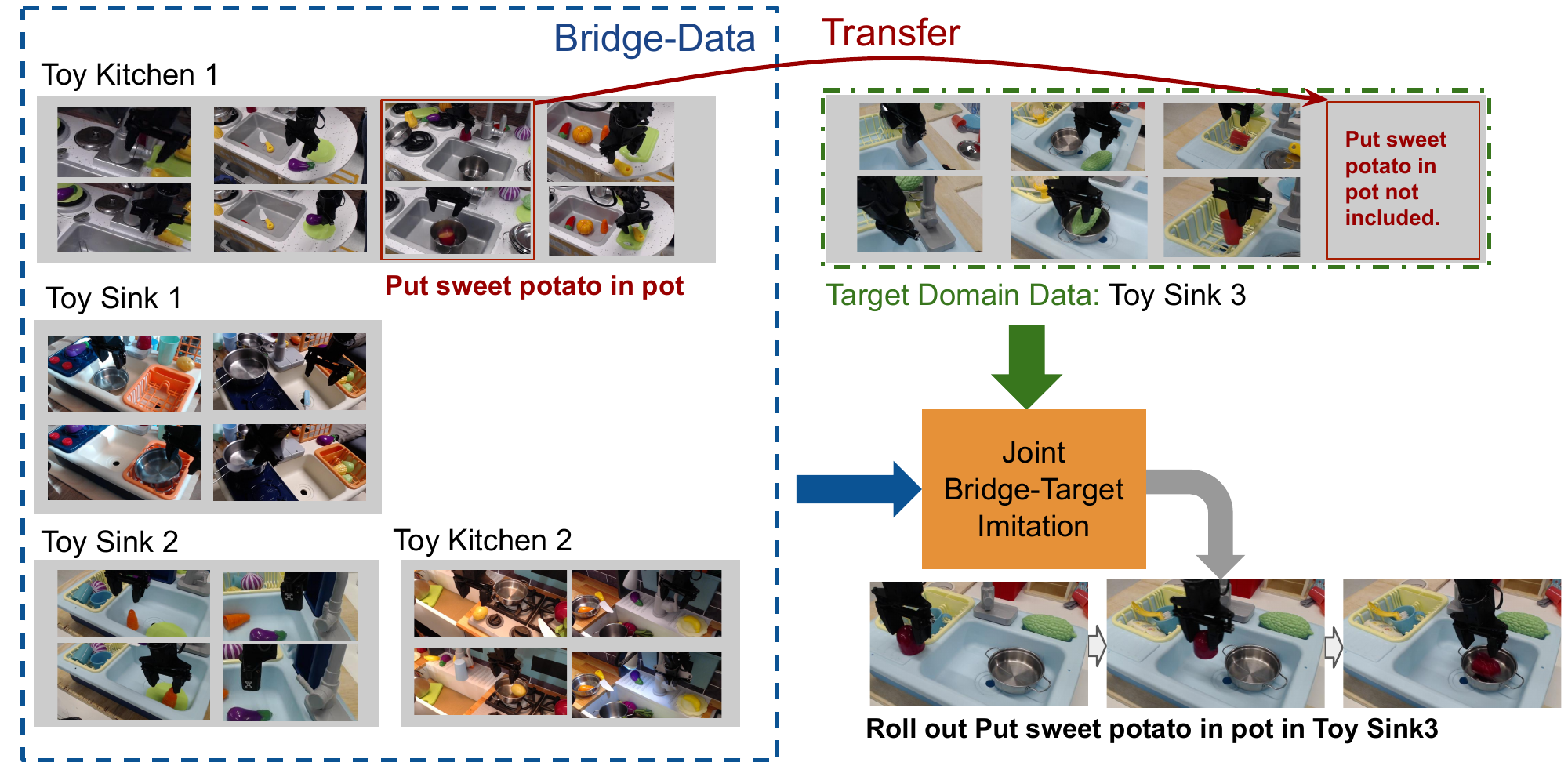}
  \caption{\textbf{Scenario (2): zero-shot transfer with target support}. In this setting, the goal is to ``import'' a task from the bridge data that was \emph{not} seen in the target domain. The user provides a few tasks in the target domain that are used to connect to the bridge data, and then asks the robot to perform a task that they did not provide, but which was seen in the bridge data. In this case, the ``put sweet potato in pot'' task is present in the toy kitchen 1 domain in the bridge data, but is \emph{not} demonstrated by the user in the target domain. After training with user-provided data for other tasks, the robot is able to perform ``put sweet potato in pot'' in the target domain.}
  \label{fig:transfer_with_targetsupport}
\end{figure}

\noindent \textbf{(2) Zero-shot transfer with target support}, where the user utilizes data from a few tasks in their target domain to ``import'' other tasks that are present in the bridge data \emph{without} additionally collecting new demonstrations for them in the target domain. For example, the bridge data contains the tasks of putting a sweet potato into a pot or a pan, the user provides data in their domain for putting brushes in pans, and the robot is then able to \emph{both} put brushes as well as put sweet potatoes in pans. We illustrate this scenario in Figure~\ref{fig:transfer_with_targetsupport}. This scenario increases the repertoires of skills that are available in the user's target environment, simply by including the bridge data, thus eliminating the need to recollect data for every task in every target environment.

\begin{figure}
\centering
  \includegraphics[width=\linewidth]{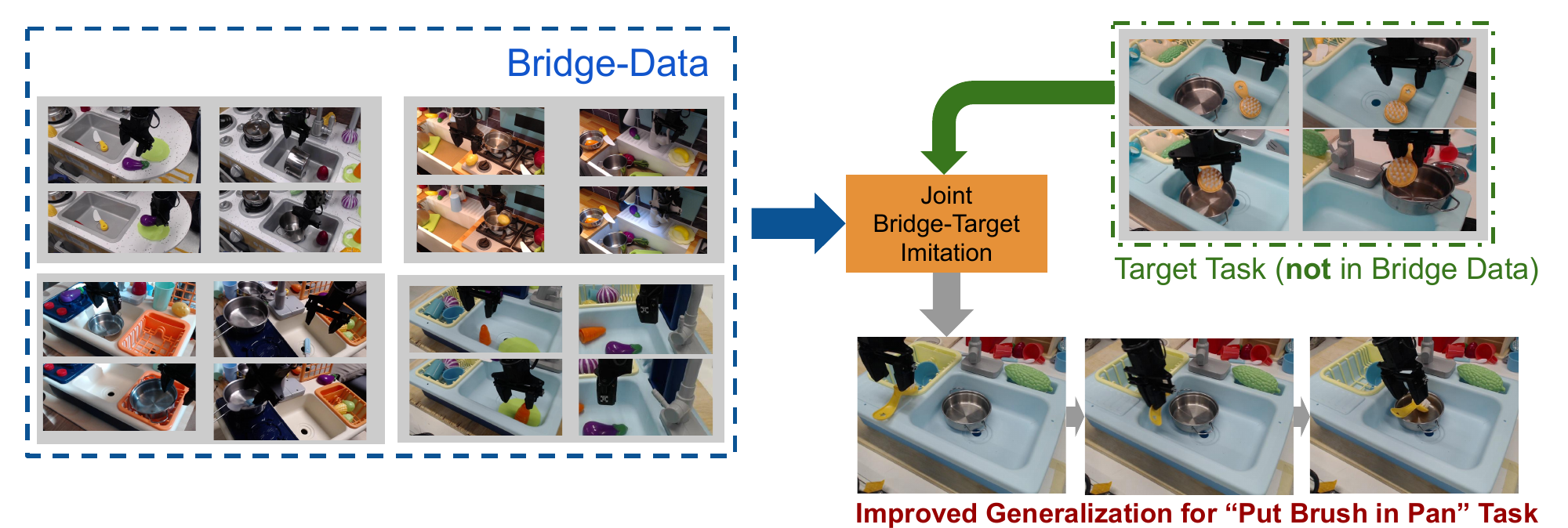}
  \caption{\textbf{Scenario (3): boosting generalization of new tasks}. The user provides some data for a \emph{new} task that was not seen in the bridge data, and the bridge data is included in training to boost performance and generalization for this new task.}
  \label{fig:novel_task_bridgedata_support}
\end{figure}

\noindent \textbf{(3) Boosting generalization of new tasks}, where the user provides a small amount of data (50 demonstrations in practice) for a new task that is not present in the bridge data, and then utilizes the bridge data to boost generalization and performance of this task. This scenario, illustrated in Figure~\ref{fig:novel_task_bridgedata_support}, most directly reflects our primary goals, since it uses the bridge data without requiring \emph{either} the domains or tasks to match, leveraging the diversity of the data and structural similarity to boost performance and generalization of entirely new tasks.

To enable this kind of generalization boosting, we conjecture that the key features that bridge datasets must have are: (i) a sufficient variety of settings, so as to provide for good generalization; (ii) shared structure between bridge data domains and target domains (i.e., it is unreasonable to expect generalization for a construction robot using bridge data of kitchen tasks); (iii) a sufficient range of tasks that breaks unwanted correlations between tasks and domains. 
Analogously to how the ImageNet dataset~\cite{deng2009imagenet} provides broad coverage that makes it possible to boost generalization for a range of computer vision tasks, the broader a bridge dataset is, the more likely target tasks receive a generalization boost in a particular target domain.

\subsection{A Bridge Dataset of Large-Scale Kitchen Tasks}
\label{sec:our_dataset}

We instantiate a bridge dataset based on the principles above as follows:

\noindent \textbf{Robotic system overview.} Since our dataset is likely the most useful for users with the same or similar type of robot, we chose to use a low-cost and widely available robot, a 6-dof WidowX250s (US\$2900), which many other users of our dataset are likely to be able to obtain. The total cost of the setup is less than US\$3600 (excluding the computer). To collect demonstrations, we use an Oculus Quest headset, where we put the headset on a table as illustrated in \autoref{fig:system_overview} next to the robot and track the user's handset while applying the user's motions to the robot end-effector via inverse kinematics. We capture images from 3 to 5 cameras concurrently, using standard webcams as well as Intel RealSense depth cameras.

\noindent \textbf{Data collection protocol.}
Our proposed bridge dataset, illustrated in \autoref{fig:dataset} consists of a total of 7200 demonstrations for 71 different tasks, collected in 10 different environments, focusing on the theme of household kitchen tasks. Each task has between 50 and 300 demonstrations. We opted to use kitchen and sink ''play sets'' for children, since they are smaller than real-world kitchens and therefore ideal for small-scale robots, and they are comparatively robust and low-cost, while still providing settings that resemble typical household scenes. 
During data collection we randomize the kitchen position (translations of 0-20cm) and the camera positions (translations of 0-10cm and rotations of 0-30 degrees) for all cameras on flexible rods every 25 trajectories. The positions of distractor objects (i.e. objects not needed for a task) are randomized at least every 5 trajectories.
All environments except toy sink 4, toy sink 5, and kitchen 3 were collected at UC Berkeley and use Logitech C920 webcams, the three remaining environments were collected at the University of Pennsylvania and use Intel RealSense RGB-D cameras. The trajectories collected at the University of Pennsylvania randomize all camera positions once every 50 trajectories.
Instructions for how users can reproduce our setup and collect data in new environments can be found on the project website.\footnote{\url{https://sites.google.com/view/bridgedata}}
\section{Using Bridge Data in Imitation Learning}

As a proof-of-concept to illustrate the utility of bridge datasets for boosting generalization in robot learning, we will present experimental results for an imitation-based approach that utilizes this data, although the data could also be used with a variety of other robotic learning algorithms such as offline RL and model-based planning.

\noindent \textbf{Incorporating bridge data.} While a variety of transfer learning methods have been proposed in the literature for combining datasets from distinct domains, we found that a simple joint training approach is effective for deriving considerable benefit from bridge data. For each of the scenarios outlined in Section~\ref{sec:boosting}, we take the user-provided demonstrations in the target domain and combine them with the entire bridge dataset for training. Since the sizes of these datasets are significantly different, we rebalance the datasets by weighting each datapoint, as discussed at the end of this section. Imitation learning then proceeds normally, simply training the policy with supervised learning on the combined dataset using the architecture described in the following paragraph. It is also possible to incorporate bridge data in other ways, for example by pretraining and finetuning. We found pretraining to be significantly less effective than joint training in our experiments, a finding that is consistent with prior works~\cite{levine2018learning}, but we emphasize that bridge datasets can be combined with target domain data in a variety of ways.
\noindent \textbf{Policy architecture.} We use task-conditioned behavioral cloning (BC) with an additional task-id input to the policy, which is used to distinguish tasks during training and testing. In some cases, a task cannot be uniquely determined by only observing the input image, and a one-hot vector representing the task will solve this issue. 
The images are first fed into a 34-layer ResNet \cite{he2016deep} and the resulting feature maps are passed through a spatial softmax \cite{finn2016deep,levine2016end}, which extracts a set of spatial positions of the relevant features. The spatial features are then concatenated with the one-hot task-id vector, and are fed into 3 layers of fully-connected networks by which the final action prediction is produced. 
During training, for a batch of training data containing tuples of task ids, images, and ground-truth actions, the network is trained by minimizing the standard $\ell_2$-error between the ground-truth actions and the predicted actions given by the policy provided the task id and the image observation as the input. 

\textbf{Training details.}
Since the amount of target domain data is usually significantly less than the amount of bridge data, we rebalance the two datasets during training. In the matching behaviors and zero-shot transfer with target support scenarios, the ratio between the number of trajectories in the bridge and target data is roughly 10:1, and we rebalance the data such that 70\% of the dataset is bridge data and 30\% is target domain data. In the ``boosting generalization of new tasks" scenario the imbalance is more severe, roughly 60:1, and so we rebalance such that 90\% of the dataset is bridge data and 10\% is target domain data. Lower rebalancing ratios of bridge data and target domain data tend to produce overfitting when the amount of target domain data is as low as 50 demonstrations.

\section{Experimental Results}
\label{sec:result}

Our experimental evaluation aims to study how well bridge data can facilitate generalization in scenarios \textbf{(1)}, \textbf{(2)}, and \textbf{(3)}, as outlined in Section~\ref{sec:boosting}. We utilize the bridge dataset described in Section~\ref{sec:our_dataset}. We evaluate generalization on a set of new target domains with limited target domain data for each of the generalization scenarios, and compare the performance of learned policies with and without bridge data.
Videos of the experiments are included in the supplementary materials and on the project webpage, which we encourage the reader to view to get a clearer sense for the diversity of the tasks: 
{\footnotesize \url{https://sites.google.com/view/bridgedata}}

\noindent \textbf{Quantitative metrics.} All quantitative evaluations use 10 trials per task,varying object positions and distractors on every trial and varying the position of the robot relative to the environment every 5 trials. This ensures that all test configurations are unique and different from any condition seen in training, providing a measurement of generalization performance for the policy. When the experiments in toy kitchen 1-3 and toy sink 1-3 were conducted, the bridge dataset only comprised 4700 trajectories. Other experiments use the full dataset with 7200 trajectories total.

\begin{figure*}[t]
\centering
    \includegraphics[width=.9\textwidth]{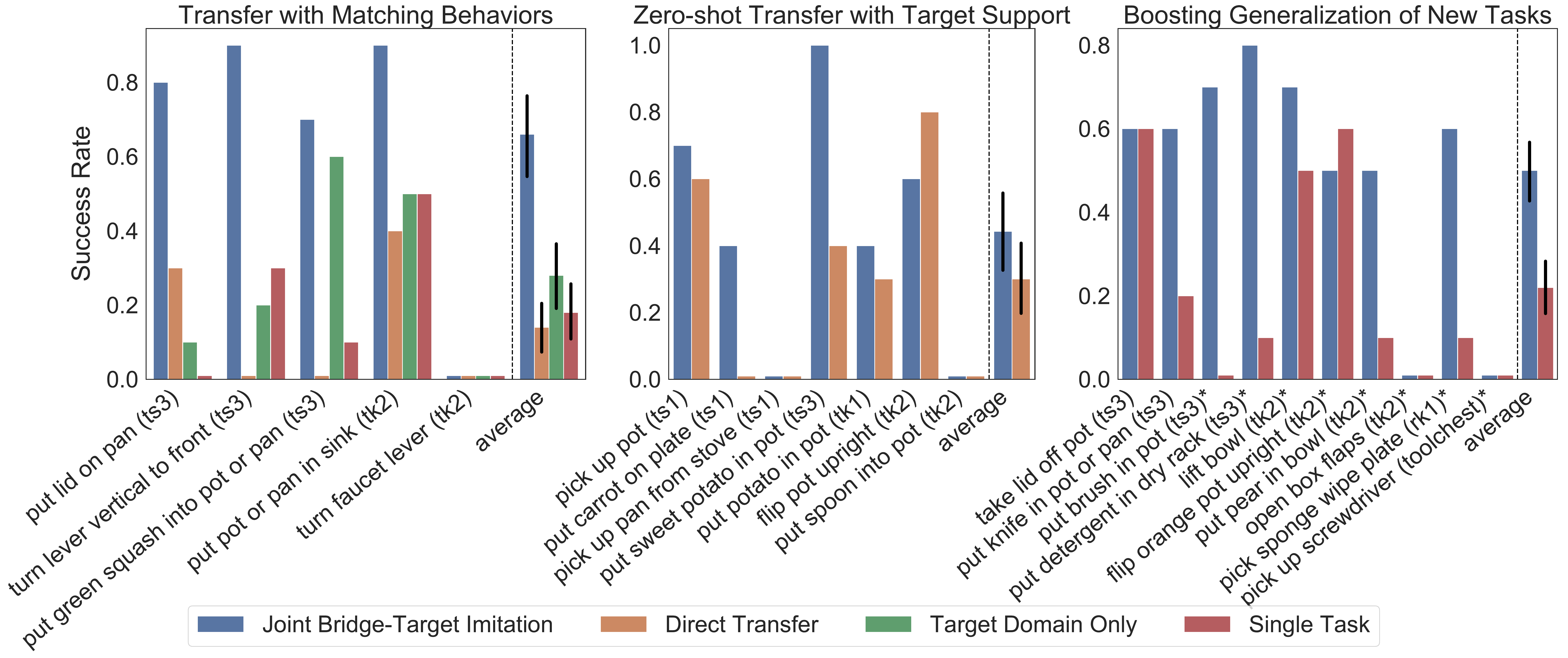}
    \caption{Comparisons of joint training with bridge data (blue) and other approaches for each type of scenario. The black vertical lines on the average success rate bar denote the standard error of the mean across different tasks for that scenario. \textbf{Left:} Performing joint training on bridge and target data leads to improved performance, here the task is included both in the bridge and target dataset. \textbf{Middle:} Using target domain data from other tasks helps transferring tasks from the bridge dataset to the target domain. \textbf{Right:} Joint training with the bridge data and a target task that is \emph{not} contained in the bridge dataset enables significant generalization improvement compared to only training on the target task alone. Tasks with an asterisk (*) uses objects that are not part of the bridge dataset.}
    \label{fig:experiment_results_bar}
    \vspace{3mm}
\end{figure*}

\begin{figure*}
\centering
    \includegraphics[width=\textwidth]{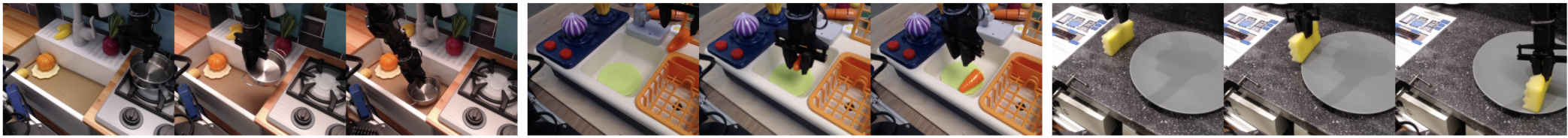}
    \caption{Examples of successful trajectories performed by the policy jointly trained with prior data and target domain data. Left: put pot in sink (scenario 1); middle: put carrot on plate (scenario 2); Right: wipe plate with sponge (scenario 3).}
    \label{fig:concat_images}
\end{figure*}

\noindent
\textbf{Scenario (1): transfer with matching behaviors.}
\autoref{fig:experiment_results_bar} (left) shows results for the transfer learning with matching behaviors scenario, where the user provides some data for a set of tasks in the target domain (which are also present in the bridge data), and we evaluate whether including bridge data during training improves performance and generalization. For comparison, we include the performance of the policy when trained \emph{only} on the target domain data, without bridge data (Target Domain Only), a baseline that uses only the bridge data without any target domain data (Direct Transfer), as well as baseline that trains a single-task policy on data in the target domain only (Single Task). The Toy Kitchen 2 (tk2) target domain has 6 tasks, and Toy Sink 3 (ts3) has 10 tasks, each with 50 demonstrations.

As can be seen in the results, jointly training with the bridge data leads to significant gains in performance ($66\%$ success averaged over tasks) compared to the direct transfer ($14\%$ success), target domain only ($28\%$ success) and the single task ($18\%$ success) baseline. This is not surprising, since this scenario directly augments the training set with additional data of the \emph{same} tasks, but it still provides a validation of the value of including bridge data in training (for a qualitative example see \autoref{fig:concat_images}, left).

%
%

\noindent
\textbf{Scenario (2): zero-shot transfer with target support.}
In the next experiment, we evaluate tasks in the target domain for which the user did \emph{not} provide any data. Instead, the user only collected data for \emph{other} tasks in the target domain. This experiment evaluates whether bridge data can be used to ``import'' tasks into the target domain. We provide a qualitative example for this scenario in \autoref{fig:concat_images} middle, which shows an experiment where we transfer the ``put carrot on plate" task into the Toy Sink 1 target domain using the bridge data and target domain data consisting of 10 \emph{other} tasks. Due to space constraints, We provide a visualization of these other tasks on the project webpage.

Since there is no target domain data for these tasks, we cannot compare to a baseline that does not use bridge data at all, since such a baseline would have no data for these tasks. However, we do include the ``direct transfer'' baseline, which utilizes a policy trained only on the bridge data. Note that this comparison is non-trivial: it is not at all clear a priori that target domain data for \emph{other} tasks should boost transfer performance of tasks that are only present in the bridge data.
The results, shown in \autoref{fig:experiment_results_bar} (middle), indicate that the jointly trained policy which obtains $44\%$ success averaged over tasks indeed attains a very significant increase in performance over direct transfer ($30\%$ success), suggesting that the zero-shot transfer with target support scenario offers a viable way for users to ``import'' tasks from the bridge dataset into their domain.

\noindent \textbf{Scenario (3): boosting generalization of new tasks.}
The last generalization scenario, which most directly evaluates the \emph{bridge data hypothesis}, aims to study how well bridge data can boost the generalization of entirely new tasks in the target domain, which are not present in the bridge data. To study this question, we collected data for 10 different unique tasks in 4 different environments and excluded them from the bridge data to simulate a user collecting their own unique task in their new target environment.
\autoref{fig:concat_images} right illustrates one of these scenarios, where we collected 50 demonstrations for the ``wipe place with sponge" task in the the real kitchen 1 target domain. \emph{Neither} data from the target domain nor this task or this object are present in the bridge data. After jointly training with both bridge and target data we obtain a significant generalization boost when running the policy in the target domain, compared to a policy trained on only the single-task target domain data. Direct transfer is impossible here, because the bridge data does not contain this task.
The results are presented in \autoref{fig:experiment_results_bar} (right), and show that training jointly with the bridge data leads to significant improvement on 6 out of 10 tasks across three evaluation environments, leading to $50\%$ success averaged over tasks, whereas single task policies attain around $22\%$ success -- a 2$\times$ improvement in overall performance (the asterisks denote in which experiments the objects are \emph{not} contained in the bridge data). 
The significant improvements obtained from including the bridge data suggest that bridge datasets can be a powerful vehicle for boosting generalization of new skills, and that a single shared bridge dataset can be utilized across a range of domains and applications. Of course, structural similarity between environments and tasks is important, and all of these evaluations use other toy kitchen or sink setups. We expect the applicability of a bridge dataset to increase as the breadth of domains and tasks in the dataset increases.
\noindent \textbf{When does bridge data help?} In \autoref{tab:bridge_data_helps} we provide a list of example scenarios where the bridge data helps and where it does not (the first 7 rows). More qualitative results, including videos of these tasks and additional discussion, are provided on the project website due to space constraints: \MYhref{https://tinyurl.com/b258nvdu}{tinyurl.com/b258nvdu}.
Qualitatively, we observed that the tasks that most consistently benefit from the inclusion of bridge data contain objects that visually resemble those seen in the bridge data (e.g., there are gains for `put pear in bowl,' where the pear resembles the vegetables in the bridge data, but no gains in `flip orange pot upright,' since the orange pot looks very different from any container in the bridge data), contain behavior that physically is related to behavior seen in the bridge data (e.g., in `put detergent in dry rack,' the bridge data helps since the motion resembles the pick-and-place motions in the prior data, whereas in `open box flaps' bridge data does not help since the type of pushing motions involved in this task are very rare in the bridge data),
and take place in domains that are visually and structurally related to those in the bridge data (e.g., the bridge data helps with `wipe plate with sponge' in \autoref{fig:concat_images} in a real kitchen, but does not help with `pick screwdriver from tool chest task,' since the scene does not resemble the toy kitchens in the bridge data).
Unfortunately, it is difficult to provide a more precise and formal treatment of when transfer learning succeeds in general, though we expect this would be an exciting direction for future research.

\section{Conclusion}
\label{sec:conclusion}
We show how a large, diverse bridge dataset can be leveraged to improve generalization in robotic learning. Our experiments demonstrate that including bridge data when training skills in a new domain can improve performance across a range of scenarios, both for tasks that are present in the bridge data and, perhaps surprisingly, entirely new tasks. This means that bridge data may provide a generic tool to improve generalization in a user's target domain. In addition, we showed that bridge data can also function as a tool to \emph{import} tasks from the prior dataset to a target domain, thus increasing the repertoires of skills a user has at their disposal in a particular target domain. This suggests that a large, shared bridge dataset, like the one we have released, could be used by different robotics researchers to boost the generalization capabilities and the number of available skills of their imitation-trained policies.  

Both our experimental evaluation and our technical approach do have a number of limitations. While we carefully set up our experiments to reflect a likely real-world usage scenario, where the target domain is distinct from the bridge data (i.e., to reflect what would happen if \emph{someone else} were to use our bridge data for their robot in their lab), we still only evaluate in a few distinct settings, namely in 5 different environments at UC Berkeley.
However, our imitation learning results do illustrate the benefits of diverse bridge data, and we hope that by releasing our dataset to the community, we can take a step toward generalizing robotic learning and make it possible for anyone to train robotic policies that readily generalize to varied environments without repeatedly collecting large and exhaustive datasets.

\begin{figure}[h]

\tiny
\setlength{\tabcolsep}{0.02in} %
\renewcommand{\arraystretch}{1.0} %

\begin{tabular}{|l|l|l|l|l|}
\hline
        \textbf{Task}  & \thead{  \textbf{ \tiny Tar.} \\ \textbf{ \tiny Env}} &  \thead{\textbf{ \tiny  Joint} \\ \textbf{ \tiny  Train}} &  \thead{ \textbf{ \tiny  Single} \\ \textbf{ \tiny Task} } & \thead{ \textbf{ \tiny Potential reason for} \\ \textbf{ \tiny no gain with bridge data}} \\ \hline

    turn faucet lever (1) & tk2 & 0\% & 0\% & \specialcell[t]{The faucet in tk2 has very different  \\ appearance from the other faucets}\\ \hline

    Pickup pan from stove (2)& ts1 & 0\% & N/A & Not enough  target domain data and prior data for this task\\ \hline

    Put spoon into pot (2) & tk2 & 0\% & N/A &  Not enough target domain data and prior data for this task \\ \hline

    flip orange pot upright (3)& tk2 & 50\% & 60\% & \specialcell[t]{There is no orange pot in prior dataset,\\ only metal pots in prior dataset}\\ \hline

    open box flaps (3) & tk2 & 10\% & 10\% & Boxes and pushing motions do not occur in  prior data\\ \hline

    take lid off pot (3)& ts3 & 60\% & 60\% & Only 100 demos  involving lids in prior dataset. \\ \hline

     pick up screw driver (3)& toolchest & 0\% & 0\% & \specialcell[t]{The toolchest and screwdrivers are \\ visually very different from prior data} \\ \hline

     put pot or pan in sink (1)& tk2 &  90\% & 50\% & \\ \hline

    put carrot on plate (2) & ts1 & 40\% & N/A  & \\ \hline    
    
    Wipe plate w/ sponge (3)& k1 & 70\% & N/A & \\ \hline
    
        put pear in bowl (3)& tk2 &  50\% & 10\% & \\ \hline
    
    put brush in pot (3)& ts3 &  90\% & 0\%  & \\ \hline
    
    put detergent dry rack (3)& ts3 &  80\% & 10\% & \\ \hline

    lift bowl (3)& tk2 &  70\% & 50\% & \\ 
    
    \hline

\end{tabular}

\caption{Comparison of scenarios where usage of the bridge data helps performance and where it does not. Scenarios where usage of bridge data does not help are marked in red font. The type of transfer setting is denoted by the number in brackets after the task description.}
\label{tab:bridge_data_helps}

\end{figure}

\bibliographystyle{./IEEEtran}
\bibliography{./ref, ./IEEEabrv}

\end{document}